\documentclass[sigconf]{acmart}
\AtBeginDocument{%
  }
\usepackage{tabularx}
\usepackage{booktabs}
\usepackage{ragged2e}
\usepackage{dirtytalk}
\newcolumntype{Y}{>{\RaggedRight\arraybackslash}X}
 
\setcopyright{acmlicensed}
\copyrightyear{2026}
\acmYear{2026}
\acmDOI{XXXXXXX.XXXXXXX}
\acmConference[FAccT '26]{ACM Conference on Fairness, Accountability, and Transparency}{June 25--26,
  2026}{Montréal, Ca}

\author{Fitsum Sileshi Beyene}
\email{fsb5110@psu.edu}
\affiliation{%
  \institution{The Pennsylvania State University}
  \city{University Park}
  \state{Pennsylvania}
  \country{USA}
}

\author{Christopher L. Dancy}
\email{cdancy@psu.edu}
\affiliation{%
  \institution{The Pennsylvania State University}
  \city{University Park}
  \country{Pennsylvania}
  \country{USA}}

\begin{document}
\title{A Survey of OCR Evaluation Methods and Metrics and the Invisibility of Historical Documents}  



\begin{abstract}

Optical character recognition (OCR) and document understanding systems increasingly rely on large vision and vision-language models, yet evaluation remains centered on modern, Western, and institutional documents. This emphasis masks system behavior in historical and marginalized archives, where layout, typography, and material degradation shape interpretation. This study examines how OCR and document understanding systems are evaluated, with particular attention to Black historical newspapers. We review OCR and document understanding papers, as well as benchmark datasets, which are published between 2006 and 2025 using the PRISMA framework. We look into how the studies report training data, benchmark design, and evaluation metrics for vision transformer and multimodal OCR systems. During the review, we found that Black newspapers and other community-produced historical documents rarely appear in reported training data or evaluation benchmarks. Most evaluations emphasize character accuracy and task success on modern layouts. They rarely capture structural failures common in historical newspapers, including column collapse, typographic errors, and hallucinated text. To put these findings into perspective, we use previous empirical studies and archival statistics from significant Black press collections to show how evaluation gaps lead to structural invisibility and representational harm. We propose that these gaps occur due to organizational (meso) and institutional (macro) behaviors and structure, shaped by benchmark incentives and data governance decisions.

\end{abstract}




\maketitle

\section{Introduction}
The design of evaluation techniques does more than just measure machine learning systems; it shapes what such systems are built to see. Indeed, connections between representations of what is acceptable according to \textit{standards} and dominant \textit{genres of the human} are often implicit in AI/ML system engineering processes \cite{WorkmanDancy2023ManAntiblackness}. In optical character recognition (OCR) and document understanding, the standards, metrics, and the (generated and managed) datasets all contribute to determining which documents are legible and whose histories get to be computed \cite{10.1145/3453476}. We argue that present OCR evaluation systems consistently fail to identify culturally relevant failure modes, essentially making entire classes of historical texts inaccessible to systems trained to read them.

Over the last decade, OCR has progressed from more character-level recognition to full integrated document understanding. Vision-language models (VLMs) and transformer-based architectures promise to extract not only the text but also semantic structure, spatial relationships, and contextual meaning from document images \cite{Xu2020LayoutLM,Huang2022LayoutLMv3,Li2023TrOCR}. However, their evaluation techniques have not kept up with this development. The dominating metrics used to evaluate these systems, such as Character Error Rate (CER) and Word Error Rate (WER), reduce document fidelity to just string-edit distance---treating document understanding as a strictly sequential transcribing activity--- which blurs the boundary between a misread character and a misread column, or a transcription error and a layout erasure \cite{10.1145/3453476}. Additionally, top benchmarks such as OCRBench, OmniDocBench, and DocVQA \cite{Liu2024OCRBench,Ouyang2025OmniDocBench,Mathew2021DocVQA} rely heavily on scientific papers, corporate forms, and current digital PDFs, which leaves historical and community archives systematically underrepresented \cite{Liu2024OCRBench,Ouyang2025OmniDocBench}. Though one may argue the advantages of escaping legibility given the problematic values implicit to the development of some VLMs, critical engagement with this part of the AI engineering life-cycle does present an opportunity to creating new critical, liberatory systems for culturally relevant digital archives.

In this review, we deliberately focus on Black digital archives, rather than attempting to cover digital archives more broadly. Beyond the important socio-technical considerations related to the ways in which these archives reflect a different set of socio-physical constraints, the early Black press digital archives we focus on are important for the current sociopolitical moment. A greater (critical, responsible) legibility and accessibility of these archives presents opportunities for use in our current moment by a wider audience (e.g., see Supreme Court Justice Ketanji Jackson's use of the Colored Conventions Project \cite{ColoredConventionsProject} in a recent dissent \cite{Jackson2024Dissent} as an important example of the power of the legible digitization of these documents). These documents include historical newspapers like Freedom's Journal (1827), The North Star (1847), and the Chicago Defender (1905), which, beyond just being interesting records of historical context, are important archives of American political, social, and cultural history. As noted by Benjamin Talton,
 \say{\textit{You can't tell the story of Black America without the Black press, because the Black press was covering things that weren't in the mainstream white newspapers and really respecting the humanity of Black people} \cite{NBCNews2024BlackPressDigitization}.} These newspapers present precisely the challenges that current evaluation systems ignore such as degraded scans from microfilm preservation, Victorian and Gothic typefaces missing from modern training corpora, and complex multi\-column layouts in which spatial arrangement carries rhetorical and political meaning \cite{BeyeneDancy2025LayoutAwareOCR,Lee2021Miseenpage,Gallon2021BlackPress}.

The consequence is a systematic blindness built into the evaluating system itself --- models deemed state-of-the-art that are trained on datasets like PubLayNet \cite{Zhong2019PubLayNet}, DocBank \cite{Li2020DocBank}, and DocLayNet \cite{Pfitzmann2022DocLayNet} show inconsistent results  when used with Black historical newspapers, partially because the evaluation systems used do not require success with these materials, which have particular sociocultural contexts that impact how their physical instantiations, digitization, and any related metadata are generated and recorded. Even if standard accuracy metrics may show satisfactory CER \cite{10.1145/3453476}, the models often silently eliminate the multi-column structure that editors used to accentuate rhetoric, promote social cohesion, and assert political identity. This error is more than just a technical problem, but also epistemological.

In this paper, we provide three contributions to the study of fairness, accountability, and transparency in document understanding:
\begin{itemize}
    \item A systematic analysis of OCR and document understanding evaluation systems, focusing on how their training data, benchmarks, and metrics are presented and implemented.
    \item An argument that current evaluation systems often fail to detect structural and layout errors common in historical newspapers, despite reporting a strong character-level accuracy and support for that argument.
    \item A case study on historical Black press materials that shows how these evaluation blind spots can result in a structurally significant errors that continue to be unreported under commonly accepted evaluation techniques.
\end{itemize}


Perhaps simpler than conversations around AI ethics, we argue that the key iterative step of evaluation for OCR systems should be revisited; though see \cite{BirhaneForgottenMargins2022,WorkmanDancy2023ManAntiblackness} for pertinent discussion around AI Ethics and the higher-level AI engineering process. The challenge is not technical per se, but rather resource allocation, dataset diversity, and recognizing cultural significance --- which may also be understood as a challenge in values \cite{BirhaneValuesEncoded2022}. By demonstrating how evaluation choices reflect assumptions about which documents matter, we hope to provide ML researchers with a vocabulary for critiquing their own benchmarks as well as a template that extends beyond Black newspapers to other important, culturally relevant digital archives such as community-supported indigenous archives and other domains where standard evaluation practices silently fail.

 \section{Related work}

This work follows from a constellation of previous research from fields such as computer vision (for OCR) and the digital humanities. We survey previous work within four interconnected research areas: the evolution of OCR systems and document understanding architectures, the development of evaluation metrics and benchmarks, studies on historical document digitization, and critical work on bias and representation in machine learning. 

\subsection{From Character Recognition to Document Understanding
}
The direction of OCR research over the last decade shows a fundamental transition from single character recognition to full document understanding. Traditional OCR systems, such as Tesseract, consider text extraction as a pipeline that comprises binarization, segmentation, and character classification, with layout analysis used as a preprocessing intermediate step \cite{SmithTesseract2007}.

Following these important systems, the LayoutLM model family represented an architectural shift by modeling text, layout, and visual aspects all within the same transformer architecture \cite{Xu2020LayoutLM}. LayoutLMv2 included cross-modal pre-training objectives that merged spatial and textual representations \cite{Xu2021LayoutLMv2}, whereas LayoutLMv3 combined text and picture masking to learn the document structure without relying on \textit{pre-extracted OCR} features \cite{Huang2022LayoutLMv3}. These models showed that treating layout as semantically important, rather than just positional, can improve the results on document understanding tasks.

Recently most of the work is towards end-to-end systems that bypass traditional OCR pipelines. For instance, TrOCR showed that transformer encoder-decoder models pre-trained on synthetic data could achieve adequate handwritten text recognition results without making other task-specific architecture alterations \cite{Li2023TrOCR}. This technique for document parsing was then extended by Donut \cite{Kim2022Donut} and Nougat \cite{Blecher2023Nougat}, which used vision transformers to generate a structured output directly from the document images. The GOT-OCR 2.0 model combined several OCR modalities, including scene text, documents, and mathematical notation, into a single architecture, yet with noticeably insufficient results on historical materials \cite{Wei2024OCR20}.

It's clear that the development of vision-language models (VLMs) represents the current edge of OCR system research. Systems like Qwen2.5-VL and PaddleOCR-VL use large-scale, multimodal pretraining to do OCR as part of the larger visual reasoning pipeline \cite{Bai2025Qwen25VL, Cui2025PaddleOCRVL}. Furthermore, the olmOCR project has focused on PDF interpretation at scale by using Internet Archive content to provide training data that includes some older texts \cite{Poznanski2025olmOCR}. However, as shown in the next section, the shift toward document comprehension has not been accompanied by equivalent changes in how we evaluate these systems.

\subsection{Evaluation Metrics: The edit distance}

Historically the evaluation of OCR systems has been based on the Character Error Rate (CER) and Word Error Rate (WER), which are both from the Levenshtein edit distance between anticipated and ground truth strings. These metrics are computationally simple and intuitive to interpret; a CER of 5\% means that one out of every twenty characters requires correction. CER/WER is almost universally used in OCR studies, with many systems reporting only these metrics \cite{10.1145/3453476, Pfitzmann2022DocLayNet}. These constraints matter because the edit-distance algorithms treat all errors the same, whether a substitution happens within a word or a line break interrupts paragraph structure. They don't have the mechanism for penalizing layout errors and if a model successfully transcribes every character but scrambles the column order it still achieves flawless CER while providing poor results. 

Even with improved metrics that begin to address some of the socially, culturally contextualized differences in historical digital archives, evaluations of various OCR systems can run into the issue of a lack of ground truth. Given the difficulty in getting many of these archives digitized, lack of such related data and metadata to provide a supervised evaluation is unsurprising. Beyene and Dancy \citeN{BeyeneDancy2025LayoutAwareOCR} introduce an unsupervised evaluation framework to potentially use for some assessments in the absence of ground-truth transcriptions of digitized archives. The authors used a combination of semantic coherence score, region entropy divergence, and textual redundancy score as a way to evaluate the performance of various OCR systems on a historical digital archive that did not have an accompanying ground truth. Though only a part of what is needed for evaluation practices that are better equipped to handle the realities of culturally relevant digital historical archives, the work does provide a useful tool given the individual, organizational, and institutional context of these digital archives (including the typical reduced resources to transcribe Black digital historical archives.)

Despite this progress, metric pluralism remains the exception instead of the standard. According to our survey, only few systems reporting OCR results contain any structural metric in addition to CER/WER, and very few particularly address layout preservation for historical texts.

\subsection{Benchmarking and Pre-Training Data}  

The benchmarks that are used to evaluate document understanding systems also set the expectations about which texts are valuable. The IIT-CDIP, which is a primary pre-training corpus for many current systems, is comprised of approximately 7 million documents, containing roughly 42 million scanned pages (images) which are collected through tobacco litigation discovery \cite{Lewis2006IITCDIP}. While IIT-CDIP provides size and diversity across document categories (letters, forms, memorandum, and reports), it is a niche institutional format (i.e., mid-to-late twentieth century American business papers) affecting the type of visual and linguistic patterns to be adopted by the models as a "standard". Modern benchmarks have increased coverage while still keeping certain gaps. OCRBench v2 has 11,500 photos from 31 contexts, including a large sum of scientific notation and scene text, but no historical document category \cite{Liu2024OCRBench}. OmniDocBench \cite{Ouyang2025OmniDocBench} offers 1,355 annotated PDFs with attribute-based evaluation of degradation factors such as blur and watermarks, however it is clearly aimed for "modern-era documents" rather than historical ones. The recently released (and still growing) olmOCR-Bench has 7,010 unit tests that emphasize layout fidelity and include some digitized print materials \cite{Poznanski2025olmOCR}, which makes it the closest to historical coverage among the major benchmarks, though it still lacks targeted representation of community or culturally relevant archives from marginalized communities.

Several benchmarks focus on document question-answering rather than OCR accuracy itself. DocVQA \cite{Mathew2021DocVQA}, DUDE \cite{VanLandeghem2023DUDE}, and DUE \cite{Borchmann2021DUE} evaluate models' ability to extract specific information from the documents, with OCR being used as an intermediary upstream component. While these standards evaluate practical utility, they can hide OCR flaws when correct responses have been generated from the incomplete or degraded text.

Recently, KITAB-Bench was created specifically for Arabic documentary heritage, such as  manuscripts and historical administrative documents, and improved by human-in-the-loop evaluation \cite{Heakl2025KITAB}. The Newspapers Navigator dataset extracted visual content from Chronicling America's historical newspaper archive, with an emphasis on image classification \cite{Lee2020NewspaperNavigator}. The American Stories research dataset provides structured article text from pre-1920 US newspapers on a large scale, showing that historical newspaper processing can be feasible \cite{Dell2023AmericanStories}.

Given the archival value of Black newspapers, their absence from evaluation benchmarks is notable. Chronicling America now includes 344 African American newspaper publications from 1777 to 1963 \cite{loc_chronicling_african_american}. While previous OCR error rates for these collections averaged 18\% \cite{burchardt_ocr_trustworthiness_2023}, the Library of Congress initiated an OCR reprocessing method in 2025 to reduce noise in early digital batches\cite{loc_ocr_reprocessing_2025}. Howard University's Black Press Archives have over 2,000 titles and 100,000 issues on 2,847 microfilm reels \cite{howard_black_press_archives}; but, as of late 2025, a major amount of these archives are still being digitized as part of a multi-year project with the goal of bringing 60\% \cite{nbc_howard_university} of the collection online \cite{crowley_digitizing_black_press}. The benchmark systems used thus reflect choices at several (individual, organizational, and institutional) levels, including institutional and financing decisions that impact which histories can become computationally legible.

\subsection{Historical Document Digitization}

Researchers studying historical document digitalization have for years recognized the challenges that mainstream OCR evaluations miss. Holley's \citeN{Holley2009HowGC} seminal work on crowd-sourcing OCR correction for Australian newspapers showed how standard methods failed on historical typography, involving human oversight at scale. More recent surveys of historical document datasets have identified the various deterioration factors such as foxing, bleed-through, uneven illumination, and physical damage that distinguishes archive materials from modern records \cite{Neudecker2021OCREvalSurvey}.

Despite these potential roadblocks, specialized systems have proven that historical OCR is technically achievable with a proper training. OCR4all achieved 84\% error reduction on 19th-century Gothic typefaces by fine-tuning on a domain-specific ground truth \cite{Reul2019OCR4all}. The fact that 80\% of German documents from 1800-1941 used Blackletter fonts and that modern systems still fail terribly on such materials without specialized training (50-70\% CER) shows how training data composition directly determines which historical periods are accessible \cite{Springmann2018}. Post-OCR correction has emerged as a complementary approach. Nguyen et al. \citeN{10.1145/3453476} ran a survey that documented solutions ranging from rule-based pattern matching to neural sequence-to-sequence models, with T5-based approaches achieving 28-40\% improvement in error reduction on historical materials \cite{Rijhwani2020EndangeredOCRError}. Beshirov et al. \citeN{Beshirov2022DuoSearch} also showed language specific post-correction for Bulgarian historical records, indicating how linguistic resources limit the historical traditions that gain from correction pipelines. This group of work establishes that there are technical capacities to process historical documents. The continuing gap in this capacity and typical system performance is due to prioritization embedded in the training dataset and the evaluation benchmarks rather than technological limitations.

\subsection{Bias, Representation, and Cultural Erasure}

In the digital humanities, researchers have explored how algorithmic systems interact with archival gaps. Gallon's \citeN{Gallon2016BlackDH} appeal for a "Black digital humanities" emphasizes the importance of accounting for the particular conditions under which Black cultural assets have been generated, preserved, and made available. The Black press, in particular, faced economic hardship and political persecution, influencing both documentation practices and archive survival conditions that typicaly deployed OCR systems have not been engineered to handle. Related recent work has started to relate these critical perspectives to technological practice. Smith et al. \citeN{Smith2013InfectiousTexts} look at text reuse in historical newspapers showing how OCR quality affects which texts can be computationally traced and which are invisible to algorithmic analysis. Casey \citeN{Casey2022BlackPress} pushed for "hemispheric" approaches to newspaper digitization with an emphasis on transnational Black press networks rather than treating publications on their own. Taurino and Smith's \citeN{Smith2022ArchivalML} definition of "machine learning as archival science" highlights that the training data curation in itself is a type of archival practice that can impact what computational systems can learn about the past. Indeed, some current (especially generative) machine learning systems are treated \textit{as} definitive forms of historical knowledge systems, continuing potentially problematic central points of information that reinforce both problematic social structures (see also \cite{Noble2018,Dancy2022AIBlackness}) and probabilistically produce false sequences of information, making the critical consideration of this \textit{archival practice} important.

Compounding the importance of understanding this archival-engineering practice within the context of machine learning with archives is the issue referred to as "over-historicization," in which VLMs generate historically appropriate but inaccurate characters, links these concerns to current model behavior. The study on GPT-4o Vision processing of 18th–19th century Russian manuscripts discovered that 59\% of errors involved anachronistic character insertions, implying that the models learn to perform "\textit{historicalness}" instead of accurately transcribing historical documents \cite{Levchenko2024HistoricalCorpora}. This failure shows a digital archival distortion that is enabled by generative AI, which can lead to the creation of plausible but false historical records.

We build on these foundations using a systematic review that connects evaluation metrics to representational outcomes. We show not only that Black newspapers are underrepresented in benchmarks, but also how this under-representation is structured by the metric choices that ignore layout, training datasets that exclude historical typography, and institutional structures that completely exclude community archives from the process of evaluation. This exclusion both hinders opportunities to gain insight into digitized archives while critically using potential OCR tools, and raises an increased possibility of encountering issues such as the previously mentioned over-historicization \cite{Levchenko2024HistoricalCorpora}.

\section{Systematic Review Methodology}

We conducted a systematic survey and review of OCR evaluation methods using PRISMA 2020 principles \cite{Page2021PRISMA}; Figure 1 shows the workflow for this survey. Our analysis ranges from the years 2006–2025, covering the transformer revolution in document understanding through the current vision-language models, and applies PRISMA to provide transparent inclusion, exclusion, and reporting criteria for evaluation-focused machine learning studies. The study is shaped by three research questions:
\begin{enumerate}
    \item On what data are state-of-the-art OCR systems trained, and how is this data reported in evaluation studies?
    \item What failure modes develop when SOTA OCR systems are applied on Black historical typography and layout?
    \item What evaluations, data, and resource shifts are needed to ensure preservation of cultural structure and accuracy?
\end{enumerate}


\subsection{Search and Selection}
We retrieved 80 papers after searching six databases: ACM Digital Library (n=15), IEEE Xplore (n=12), arXiv (n=18), ACL Anthology (n=10), Springer (n=8), and Other (Google Scholar, Semantic Scholar, etc.) (n=17). We classified the search queries used into four categories: primary evaluation (e.g., "OCR evaluation metrics document understanding," "document AI benchmark evaluation"), historical and bias-aware (e.g., "historical document OCR evaluation," "newspaper digitization OCR evaluation"), model-specific (e.g., "vision transformer OCR evaluation"), and negative control (validation); see Appendix A for a  full list of the search queries used. After removing 2 duplicates, 78 unique records remained for screening.

Following screening, the inclusion criteria was applied to keep papers which (1) presented or evaluated OCR systems using quantitative metrics as a key contribution, (2) introduced or analyzed evaluation benchmarks, or (3) provided an overview of OCR evaluation processes. 

\subsection{Data Extraction}

For each system, we extracted the model architecture, training data sources, data size and transparency, language coverage, and historical document coverage. Table 1 shows this analysis for 13 major systems, showing the variation in reporting transparency that many systems only describe training data as "undisclosed" or "web-scale." For each benchmarks, we recorded supported languages, document types, historical coverage, degradation testing, and whether underrepresented or community archives were explicitly included. Table 3 summarizes the major benchmarks and identifies gaps between benchmark scope and the requirements of historical archives.

 \begin{figure}[h]
  \centering
  \includegraphics[width=\columnwidth]{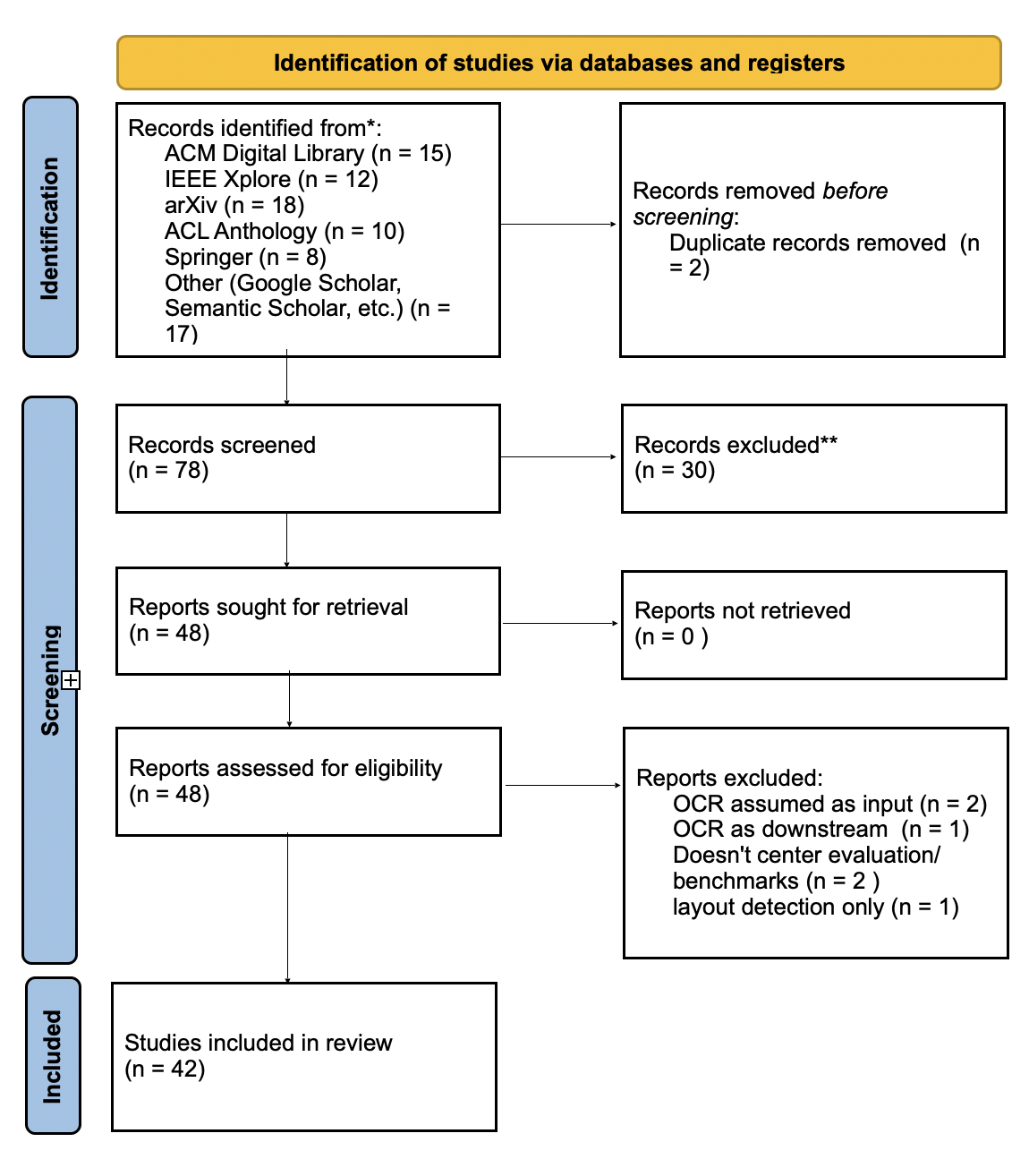}
  \Description{PRISMA 2020 flow diagram}
  \caption{PRISMA 2020 flow diagram for the systematic review of OCR evaluation methods and metrics in document understanding (2006--2025).}
  \label{fig:prisma}
\end{figure}
\section{Systematic Review Results}
\subsection{What Current OCR Evaluation Measures and What It Misses}
Table \ref{tab:training-data-alt} provides the training sources for the 13 OCR systems where six of the models rely on the IIT-CDIP database of 42 million tobacco litigation documents from the mid-twentieth century. The rest use synthetic data, scientific PDFs (PubMed Central), or web-scale corpora of varying composition. OlmOCR 2 uses a specialized data mix known as olmOCR-mix-1025, that combines broad data from the Internet Archive and S2PDF with a set of 20,000 historical and handwritten documents. The model is optimized to handle the complex and degraded scans found in archival materials, such as 19th-century correspondence from the Library of Congress, using Reinforcement Learning with Verifiable Rewards (RLVR). The training data reported includes scientific publications (360K+), corporate and legal documents (42M pages), and modern digital PDFs. From this distribution of data sources, it's clear that Historical Black newspapers, community publications, Gothic and Blackletter typefaces, and non-Western document layouts are examples of missing or unreported model training data.
\begin{table}[h]
\captionsetup{justification=raggedright, singlelinecheck=false}
\caption{Training Data Sources and Historical Coverage of \\ Major OCR Systems (2021--2025)}
\label{tab:training-data-alt}
 
\scriptsize
\setlength{\tabcolsep}{5pt}
\begin{tabular}{@{}p{1.1cm}p{0.3cm}p{1.8cm}p{1.0cm}p{0.7cm}p{0.5cm}p{1.7cm}@{}}
\toprule
\textbf{Model} & \textbf{Year} & \textbf{Training Source} & \textbf{Scale} & \textbf{Params} & \textbf{Scripts} & \textbf{Historical Coverage} \\
\midrule
olmOCR 2 & 2025 & Web PDFs \& Internet Archive & \texttildelow260K pgs & \texttildelow7B & EN+ & \textbf{Substantial} (IA scans; no curation) \\[3pt]
PaddleOCR-VL & 2025 & Synthetic \& web corpora & Undisclosed & \texttildelow0.9B & 109+ & Limited (no historical eval) \\[3pt]
Qwen2.5-VL & 2025 & Web multimodal corpora & Web-scale & 7B/72B & Multi & None reported \\[3pt]
GOT-OCR 2.0 & 2024 & Synthetic + modern web & \texttildelow5M pairs & \texttildelow580M & EN & None (synthetic focus) \\[3pt]
Surya & 2024 & Scanned document images & Undisclosed & Undiscl. & 90+ & Limited (no benchmarks) \\[3pt]
DocFormer v2 & 2024 & Document imgs (IIT-CDIP) & \texttildelow11M imgs & 200--500M & EN & None reported \\[3pt]
Nougat & 2023 & PubMed Central PDFs & \texttildelow1M docs & \texttildelow350M & EN & None (academic PDFs) \\[3pt]
UDoc & 2021 & IIT-CDIP regions & Undisclosed & Undiscl. & EN & None \\[3pt]
LayoutLMv3 & 2022 & IIT-CDIP & \texttildelow11M docs & 133/368M & EN & Partial (litigation docs) \\[3pt]
DiT & 2022 & IIT-CDIP & \texttildelow42M imgs & 33--675M & EN & Partial (structure focus) \\[3pt]
Donut & 2022 & IIT-CDIP + SynthDoG & \texttildelow13M imgs & \texttildelow200M & 4 langs & None (synthetic) \\[3pt]
TrOCR & 2021 & Synthetic + IAM & Undisclosed & \texttildelow558M & EN & Fine-tuning only \\[3pt]
Tesseract v5 & 2021 & Synthetic fonts & Varies & N/A & 100+ & Custom training required \\
\bottomrule
\end{tabular}
\vspace{3pt}
\raggedright
 
\end{table}
\\
 Table \ref{tab:ocr-evaluation} shows uneven evaluation coverage as only three systems report quantitative results for historical scans. For example, olmOCR 2 reports 82.3\% accuracy on old math scans, but this drops to 47.7\% for general historical scans. In contrast, GOT-OCR 2.0 achieves only 22.1\% on the same olmOCR-Bench, while TrOCR reports 5.7\% CER even after fine-tuning. Most historical typography entries are "Not evaluated" or "Requires fine-tuning," whereas modern capabilities are more thoroughly documented.

\begin{table}[h]
 \captionsetup{justification=raggedright, singlelinecheck=false}
\caption{Comparative Evaluation of Critical Document Analysis Systems (2025)}
\label{tab:ocr-evaluation}
\setlength{\tabcolsep}{4pt}
\renewcommand{\arraystretch}{1.1}
\scriptsize
\begin{tabular}{@{}p{1.0cm}p{1.0cm}p{2.0cm}p{1.0cm}p{1.2cm}p{1.3cm}@{}}

\toprule
\textbf{System} & \textbf{Architecture} & \textbf{Historical Scans} & \textbf{Historical Typography} & \textbf{Complex Layouts} & \textbf{Handwriting} \\
\midrule
\multicolumn{6}{c}{\textit{Generative / VLM (End-to-End Text Generation)}} \\
\midrule
\textbf{olmOCR 2} & VLM (7B) & \textbf{82.3\%} (Math)\newline 47.7\% (General) & Supported & 83.7\% (Multi-col) & Supported (20k pages) \\
\textbf{Qwen2.5-VL} & VLM (7B+) & 65.5\% (Bench) & High (Gen. Cap.) & \textbf{Strong} & Remarkable (Script-agnostic) \\
\textbf{GOT-OCR 2.0} & Encoder-Dec. & 22.1\% (Fail on noise) & Poor & Poor (reported) & Not evaluated \\
\textbf{Nougat} & Transformer & Poor (Hallucinates) & N/A & Academic Only & Poor \\
\textbf{Donut} & Transformer & Poor & Poor & Poor & Weak \\
\textbf{TrOCR (L)} & Transformer & 5--7\% CER (FT) & Fine-tunable & N/A (Line-level) & \textbf{2.89\% CER} (IAM) \\

\midrule
\multicolumn{6}{c}{\textit{Discriminative / Hybrid (Detection \& Recognition)}} \\
\midrule
\textbf{Surya} & Hybrid (Seg.) & 81\% (Hist. Tamil)\textsuperscript{\textdagger} & Moderate & Moderate & Not evaluated \\
\textbf{PaddleOCR v5} & 2-Stage & Moderate (Modern) & Moderate & Good (Tables) & High (82--89\% Acc.) \\
\textbf{Tesseract v5} & LSTM & High WER (Degraded) & Requires Fine-tune & Poor & Poor \\

\midrule
\multicolumn{6}{c}{\textit{Visual Document Understanding (Requires External OCR)}} \\
\midrule
\textbf{DocFormer v2} & Multimodal & \textit{Dep. on Input OCR} & \textit{Dep. on Input OCR} & Moderate & N/A \\
\textbf{LayoutLMv3} & Multimodal & \textit{Dep. on Input OCR} & \textit{Dep. on Input OCR} & \textbf{SOTA} & N/A \\
\textbf{UDoc / DiT} & Image Trans. & \textit{Dep. on Input OCR} & \textit{Dep. on Input OCR} & Strong & N/A \\

\bottomrule
\end{tabular}
\vspace{3pt}
\raggedright
\textit{\textdagger "Dep. on Input OCR" indicates models that classify layout but do not generate text themselves.}
\end{table}
Table 3 summarizes seven benchmarks, which are organized by language coverage, document type, historical inclusion, and degradation testing, highlighting their limited engagement with historical and community-produced documents. CC-OCR (ICCV 2025), for instance, supports over 28 languages and focuses on modern applications, while OmniDocBench (CVPR 2025) evaluates blur, rotation, and watermarks without archival focus. KITAB-Bench is unique in that it explicitly includes historical Arabic documents with human-in-the-loop review.

\begin{table}[h]
 \captionsetup{justification=raggedright, singlelinecheck=false}
\caption{Evaluation Coverage of Major OCR and Document Understanding Benchmarks}
\label{tab:benchmark-coverage}
\setlength{\tabcolsep}{3pt} 
\renewcommand{\arraystretch}{1.1} 
\scriptsize
\begin{tabular}{@{}p{1.1cm}p{1.0cm}p{1.3cm}p{1.5cm}p{1.6cm}p{1.5cm}@{}}
\toprule
\textbf{Benchmark} & \textbf{Languages} & \textbf{Document Types} & \textbf{Annotation Paradigm} & \textbf{Hist. / Archival Coverage} & \textbf{Underrep. Archives} \\
\midrule
\textbf{CC-OCR} & Multi (28+) & Web PDF, scene, docs & Plain Text & None (modern web-crawl) & No targeted focus \\
\addlinespace[3pt]
\textbf{OmniDoc} & EN, ZH & Academic, slides, books & Detection \& Text & Modern-era (Born-digital) & No targeted focus \\
\addlinespace[3pt]
\textbf{OCRBench} & EN, ZH & 31 scenarios (mixed) & VQA / Key-Value & Limited (Isolated IAM/HKR) & No explicit focus \\
\addlinespace[3pt]
\textbf{DUDE} & Prim. EN & Multi-page business & QA Pairs & Contemporary (Scanned) & No targeted focus \\
\addlinespace[3pt]
\textbf{DUE} & English & Tables, graphs & QA / Classif. & None & No targeted focus \\
\addlinespace[3pt]
\textbf{DocVQA} & English & Industry forms (Tobacco) & QA Pairs (ANLS) & Legacy Litigation (20th C.) & Single-domain \\
\addlinespace[3pt]
\textbf{KITAB*} & Arabic & Manuscripts, Chronicles & Transcription & \textbf{High} (Primary Focus) & \textbf{Arabic Heritage} \\
\bottomrule
\end{tabular}
\vspace{3pt}
\raggedright
\textit{*Refers to OpenITI/KITAB project evaluation subsets for classical Arabic texts.}
\end{table}

 
\subsection{Black Press as a Case Study}
To further test the gaps in OCR system development for culturally relevant digitized historical archives, we compared three of the models found in our review to parse a page from The Weekly Advocate (1837), one of the first African American newspapers published in New York City. Here, we use this case study to go beyond aggregate benchmark scores and observe how evaluation gaps result in specific issues with how the archive may be automatically transcribed and interpreted by these OCR systems. We tested three systems from those listed in Table \ref{tab:ocr-evaluation}: Tesseract v5 (a \textit{traditional hybrid pipeline}), Surya (a \textit{layout-centric modern system}), and olmOCR 2 (a \textit{state-of-the-art VLM}). We observed a different failure pattern for each system through these tests, which are described in Table \ref{tab:error-regimes}. 
\begin{table}[h]
 \raggedright
 
 \captionsetup{justification=raggedright, singlelinecheck=false}
\caption{Failure patterns from the \textit{The Weekly Advocate} (1837)}
\label{tab:error-regimes}
\setlength{\tabcolsep}{4pt}
\renewcommand{\arraystretch}{1.0}
\scriptsize
\begin{tabular}{@{}p{1.2cm}p{1.1cm}p{2.7cm}p{2.7cm}@{}}
\toprule
\textbf{Error Regime} & \textbf{Model} & \textbf{Technical Failure} & \textbf{Archival Consequence}  \\
\midrule
\textbf{Geometric Collapse} \newline (Linearity Bias) & Tesseract v5 & Misinterpreted vertical column rules as text; defaulted to left-to-right ``Z-pattern'' reading order & Semantic merging of distinct content (editorial poetry conflated with civic reports); granular search rendered ineffective \\
\addlinespace[4pt]
\textbf{Decoding Instability} & Surya & High-contrast 19th-c. fonts triggered token collapse; generative repetition of character sequences & Introduction of ``garbage tokens'' \\
\addlinespace[4pt]
\textbf{Corrective Hallucination} & olmOCR 2 \newline (SOTA VLM) & Overwrote a visual evidence with high-probability tokens (``knowledge bleed?'') & Entities were replaced with a fabricated text;   \\
\bottomrule
\end{tabular}
\end{table}

\begin{table*}[h]
\centering
\captionsetup{justification=raggedright,singlelinecheck=false}
\caption{Structural, temporal, and archival characteristics of some Black press documents.}
\label{tab:dataset-card}
\setlength{\tabcolsep}{3pt}
\renewcommand{\arraystretch}{1.15}
\scriptsize
\begin{tabularx}{\textwidth}{p{2.6cm} p{1.5cm} Y Y Y Y Y p{1.2cm}}
\toprule
\textbf{Publication / Series} &
\textbf{Years} &
\textbf{Layout Topology} &
\textbf{Masthead / Header} &
\textbf{Non-Text Elements} &
\textbf{Typography Complexity} &
\textbf{Scan Degradation}  
  \\
\midrule

\textbf{The Weekly Advocate \& The Colored American} (Transitional newspapers) &
1837--1841 &
4 columns, standard grid frequently interrupted by advertisements \& spanning headers &
High complexity, ribbon banners with multiple embedded text (e.g., ``Established for \ldots'' \& bold serif headers with sub-text &
High variance, heavy horizontal rules, woodcut engraving of the U.S.\ Capitol, \& iconographic glyphs in advertisements &
Moderate--high, modern serif (Didone), tight leading, archaic ligatures & bleed-through \\

\addlinespace[3pt]

\textbf{Mirror of Liberty} (Quarterly magazine) &
1838--1841 &
2 columns, wider column width, magazine aspect ratio &
Low complexity, simple typographic header, outline block type &
Minimal, primarily text-heavy, sparse rules &
Moderate, larger point size, clear typographic hierarchy &
low contrast, character dropout  
  \\

\addlinespace[3pt]

\textbf{The North Star} (Abolitionist) &
1847--1851 &
7 columns, extremely dense layout &
Moderate, typographic masthead often obscured by scan artifacts &
Structural emphasis, thin and broken vertical rules &
Very high, minimal point size, dense setting, italics for emphasis &
blur, speckle noise, skew \\

\addlinespace[3pt]

\textbf{The Impartial Citizen} &
1861--1864 &
5 columns, dense grid with slightly wider margins &
High, Gothic/Blackletter masthead distinct from Roman body text &
Standard, vertical column rules &
High, dense text blocks challenge line segmentation &
lighting gradients from microfilm \\

\addlinespace[3pt]

\textbf{The Weekly Anglo-African} &
1859--1861 &
7 columns, header-centered structure &
Moderate, detailed masthead with integrated text &
Integrated artwork, motto embedded in engraving &
Mixed, display fonts in header vs.\ standard body text &
cleaner lines than other samples  
\\

\bottomrule
\end{tabularx}
\end{table*}

\begin{figure}[h]
    \raggedright
    \captionsetup{justification=raggedright, singlelinecheck=false}
    \caption{Original vs Surya Layout analysis vs Tesseract}
    \includegraphics[width=0.8\linewidth]{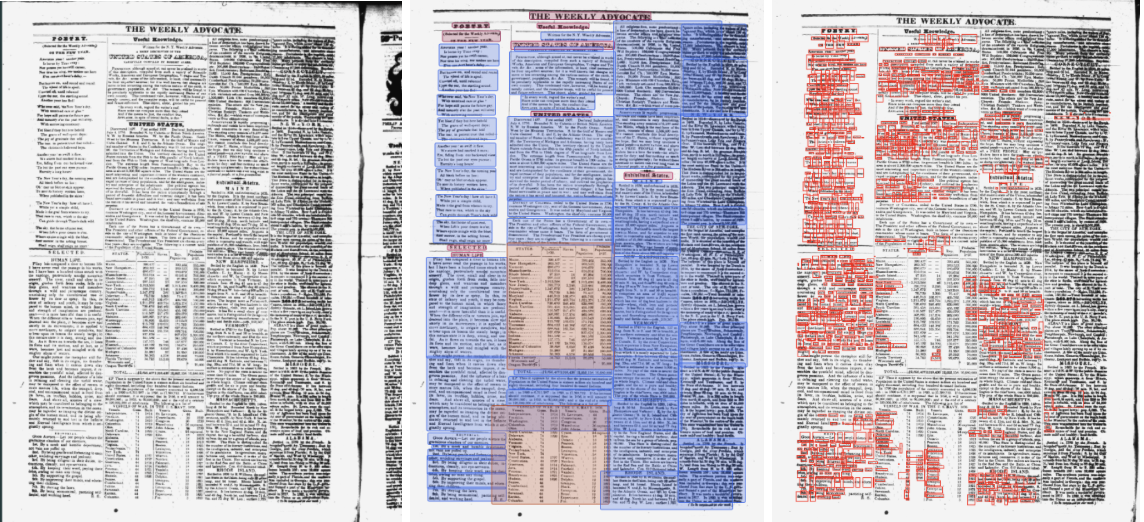}
    
    \label{fig:placeholder}
   
\end{figure}

Our analyses show a significant gap between the performance of the reported state-of-the-art (SOTA) OCR systems and the usefulness of these systems for historical archival documents that represent the social, cultural, and physical realities shown with those such as historical Black newspapers. We show that the "metric pluralism" needed to capture the accuracy of historical documents is largely missing from current evaluation practice. Although layout-aware models and task-specific layout evaluations exist, they are typically trained on web-scale or corporate collections, such as IIT-CDIP, and then compared to benchmarks that prioritize modern layouts and visually obvious degradations over historical structure and reading order. Our case study shows that while models report satisfactory character recognition scores, they often produce hallucinated content or fail to retain the reading order of multi-column Black newspapers. Current evaluation methods, which fail to identify representational harm, allow models to be deemed "successful" while ignoring the social and physical constraints shaped by historical Black print culture.

\section{Discussion}

\subsection{Evaluation gap}

Currently available evaluation pipelines suffer from algorithmic invisibility, where failure modes particular to historically underrepresented documents are not reported by existing benchmarks. Even though Chronicling America explicitly focuses on digital archives in the United States, it still remains underutilized in favor of benchmarks such as DocVQA. Our case study supports this observation by evaluating a newspaper from the Black Digital Archives, The Weekly Advocate, which displays varied layouts and typography and how the failures exhibited by the OCR models remains overlooked. Therefore, without deliberate inclusion of such documents for evaluation systems and training loops, these systems will continue to regard Black cultural heritage as an outlier rather than a core domain.

\subsection{Synthetic vs Real Dataset}

There is a major difference between a synthetic data augmentation and a historical degradation. It is very common to see OCR models handling synthetic perturbations such as Gaussian noise or artificial speckle really well. In contrast, the Early Black Press is characterized by materially specific artifacts including micro-film lighting gradients, physical warping, and irreversible binarization loss. As shown from our case study, models that seem robust under synthetic noise often collapse under such structural degradations by creating a misleading sense of reliability that can not be applied to the real-world archival digitization.

\subsection{Technical accuracy vs Preserving culture \& history}

We observed that the standard OCR criteria can create an unresolved tension between optimizing for textual "accuracy" or "cleanliness" and preserving the historical documents as cultural archives. This historical document's layout contain their editorial logic through complex, multi-column designs that maximize the scarce print space usage and can convey additional meaning \cite{Lee2021Miseenpage}. SOTA systems then enforce linear reading orders or try to reduce "noise" like marginalia to improve character level accuracy while erasing the sematic structure of the papers. Therefore, future benchmarks must consider beyond Word Error Rate (WER) to include layout-aware and fidelity oriented evaluation criteria that will be able to respect the integrity of the archival record.

\section{Limitations and Future Work}

Our study used a small, particular case study aiming to highlight the structural diversity, rather than statistical accuracy. We focused on a qualitative failure analysis instead of ground-truth transcriptions, which also reflects the constraints of historical archives limiting the direct comparison using character level accuracy metrics. We also focused on nineteenth-century Black newspapers in the United States and their specific degradations and typographic patterns, which may not generalize to other historical or linguistic contexts. Finally, our evaluation prioritized archival implications above task specific performance, which leaves the development of layout and fidelity-aware metrics to future work.

Additionally, there remains a critical question of the best ways to encourage engagement with such important historical archives. Though it is important and indeed useful to be able to understand both the content of these archives and the implications of this content (see \cite{Jackson2024Dissent}), the question of how accessible to make such archives in the face of possible problematic use of those archives remains, including potential issues such as over-historicization \cite{Levchenko2024HistoricalCorpora}. As we navigate these potential critical questions for various Black Liberatory collections \cite{Croskey2025LibCollections}, we hope to keep in mind the ways meaning has been communicated across time with a mind for the ways in which meaning was purposefully obscured \cite{Morrison2022BlackComputationalThought} and how one may continue such traditions in this work (and the limitations thereof).

\section{Conclusion}  

A major challenge to effective OCR systems for the Early Black Press, moves beyond technical capability, to questions of resource allocation, dataset diversity, and recognition of cultural significance. Though modern systems can parse complex historical layouts, they are rarely evaluated against such varied typographic and document conditions of Black newspapers, which leads to the erasure of editorial logic, and questions community priorities. This paper argues for evaluation practices to treat Black digital archives not as edge cases, but as a part of the main test dataset for a critical, culturally responsible OCR evaluation and benchmark systems. Though there are limitations to inclusion (which in of itself can be predatory \cite{Taylor2019RaceProfit,Dancy2022AIBlackness}), such a change could prove important for key understanding of such historical archives towards different futures.

\section{Generative AI usage statement}
Though generative AI systems are part of the subject of this article (for OCR systems), the authors did not make any use of generative AI to write this paper in any way of which they are aware.
\newpage

\bibliographystyle{ACM-Reference-Format}
\bibliography{ref}

@inproceedings{WorkmanDancy2023ManAntiblackness,
author    = {Deja Workman and Christopher L. Dancy},
title     = {Identifying Potential Inlets of Man in the Artificial Intelligence Development Process: Man and Antiblackness in AI Development},
booktitle = {Companion Publication of the 2023 Conference on Computer Supported Cooperative Work and Social Computing},
series    = {CSCW '23 Companion},
year      = {2023},
isbn      = {9798400701290},
pages     = {348--353},
numpages  = {6},
publisher = {Association for Computing Machinery},
address   = {New York, NY, USA},
doi       = {10.1145/3584931.3606981},
url       = {https://doi.org/10.1145/3584931.3606981}
}

@article{10.1145/3453476,
author = {Nguyen, Thi Tuyet Hai and Jatowt, Adam and Coustaty, Mickael and Doucet, Antoine},
title = {Survey of Post-OCR Processing Approaches},
year = {2021},
issue_date = {July 2022},
publisher = {Association for Computing Machinery},
address = {New York, NY, USA},
volume = {54},
number = {6},
issn = {0360-0300},
url = {https://doi.org/10.1145/3453476},
doi = {10.1145/3453476},
journal = {ACM Comput. Surv.},
month = jul,
articleno = {124},
numpages = {37}
}

@inproceedings{Xu2020LayoutLM,
author    = {Yiheng Xu and Minghao Li and Lei Cui and Shaohan Huang and Furu Wei and Ming Zhou},
title     = {LayoutLM: Pre-training of Text and Layout for Document Image Understanding},
booktitle = {Proceedings of the 26th ACM SIGKDD International Conference on Knowledge Discovery \& Data Mining},
series    = {KDD '20},
year      = {2020},
isbn      = {9781450379984},
pages     = {1192--1200},
numpages  = {9},
publisher = {Association for Computing Machinery},
address   = {New York, NY, USA},
doi       = {10.1145/3394486.3403172},
url       = {https://doi.org/10.1145/3394486.3403172}
}

@inproceedings{Huang2022LayoutLMv3,
author    = {Yupan Huang and Tengchao Lv and Lei Cui and Yutong Lu and Furu Wei},
title     = {LayoutLMv3: Pre-training for Document {AI} with Unified Text and Image Masking},
booktitle = {Proceedings of the 30th ACM International Conference on Multimedia},
series    = {MM '22},
year      = {2022},
isbn      = {9781450392037},
pages     = {4083--4091},
numpages  = {9},
publisher = {Association for Computing Machinery},
address   = {New York, NY, USA},
doi       = {10.1145/3503161.3548112},
url       = {https://doi.org/10.1145/3503161.3548112}
}

@inproceedings{Li2023TrOCR,
author    = {Minghao Li and Tengchao Lv and Jingye Chen and Lei Cui and Yijuan Lu and Dinei Florencio and Cha Zhang and Zhoujun Li and Furu Wei},
title     = {{TrOCR}: Transformer-Based Optical Character Recognition with Pre-trained Models},
booktitle = {Proceedings of the Thirty-Seventh AAAI Conference on Artificial Intelligence},
series    = {AAAI'23},
year      = {2023},
isbn      = {978-1-57735-880-0},
articleno = {1469},
numpages  = {9},
publisher = {AAAI Press},
doi       = {10.1609/aaai.v37i11.26538},
url       = {https://doi.org/10.1609/aaai.v37i11.26538}
}

@article{Liu2024OCRBench,
author    = {Yuhang Liu and Zhe Li and Minghao Huang and others},
title     = {{OCRBench}: On the Hidden Mystery of {OCR} in Large Multimodal Models},
journal   = {Science China Information Sciences},
volume    = {67},
articleno = {220102},
year      = {2024},
publisher = {Springer},
doi       = {10.1007/s11432-024-4235-6},
url       = {https://doi.org/10.1007/s11432-024-4235-6}
}

@inproceedings{Ouyang2025OmniDocBench,
author    = {Linke Ouyang and Yuan Qu and Hongbin Zhou and Jiawei Zhu and Rui Zhang and Qunshu Lin and Bin Wang and Zhiyuan Zhao and Man Jiang and Xiaomeng Zhao and Jin Shi and Fan Wu and Pei Chu and Minghao Liu and Zhenxiang Li and Chao Xu and Bo Zhang and Botian Shi and Zhongying Tu and Conghui He},
title     = {{OmniDocBench}: Benchmarking Diverse {PDF} Document Parsing with Comprehensive Annotations},
booktitle = {Proceedings of the IEEE/CVF Conference on Computer Vision and Pattern Recognition},
series    = {CVPR 2025},
year      = {2025},
pages     = {24838--24848},
publisher = {IEEE},
doi       = {10.1109/CVPR52734.2025.02313},
url       = {https://doi.org/10.1109/CVPR52734.2025.02313}
}

@inproceedings{Mathew2021DocVQA,
author    = {Minesh Mathew and Dimosthenis Karatzas and C. V. Jawahar},
title     = {{DocVQA}: A Dataset for {VQA} on Document Images},
booktitle = {Proceedings of the IEEE Winter Conference on Applications of Computer Vision},
series    = {WACV 2021},
year      = {2021},
pages     = {2199--2208},
publisher = {IEEE},
doi       = {10.1109/WACV48630.2021.00225},
url       = {https://doi.org/10.1109/WACV48630.2021.00225}
}

@misc{ColoredConventionsProject,
author       = {{Colored Conventions Project}},
title        = {About the Colored Conventions Project},
url          = {https://omeka.coloredconventions.org/about},
note         = {Retrieved January 14, 2026}
}

@misc{Jackson2024Dissent,
author       = {{Jackson, Ketanji Brown}},
title        = {Dissenting Opinion in \textit{Case No. 23--1275}},
year         = {2024},
howpublished = {Supreme Court of the United States},
url          = {https://www.supremecourt.gov/opinions/24pdf/23-1275_e2pg.pdf},
note         = {Dissenting opinion}
}

@inproceedings{BeyeneDancy2025LayoutAwareOCR,
author    = {Fitsum Sileshi Beyene and Christopher L. Dancy},
title     = {Layout-Aware {OCR} for Black Digital Archives with Unsupervised Evaluation},
booktitle = {Proceedings of the IEEE International Symposium on Technology and Society},
series    = {ISTAS 2025},
year      = {2025},
pages     = {1--5},
publisher = {IEEE},
doi       = {10.1109/ISTAS65609.2025.11269615},
url       = {https://doi.org/10.1109/ISTAS65609.2025.11269615}
}

@misc{Gallon2021BlackPress,
author       = {Kim Gallon},
title        = {The Black Press},
year         = {2021},
month        = sep,
journal      = {Oxford Research Encyclopedia of American History},
publisher    = {Oxford University Press},
url          = {https://oxfordre.com/americanhistory/view/10.1093/acrefore/9780199329175.001.0001/acrefore-9780199329175-e-851},
note         = {Retrieved January 14, 2026}
}

@misc{NBCNews2024BlackPressDigitization,
author       = {Curtis M. Bunn},
title        = {Howard University Is Digitizing Thousands of Black Newspapers},
year         = {2024},
journal      = {NBC News},
url          = {https://www.nbcnews.com/news/nbcblk/howard-university-digitize-archive-thousands-black-newspapers-rcna17185},
note         = {Retrieved January 14, 2026}
}

@inproceedings{Zhong2019PubLayNet,
author    = {Xingjiao Zhong and Jianbin Tang and Antonio Jimeno Yepes},
title     = {PubLayNet: Largest Dataset Ever for Document Layout Analysis},
booktitle = {Proceedings of the 15th International Conference on Document Analysis and Recognition},
series    = {ICDAR 2019},
year      = {2019},
publisher = {IEEE},
url       = {https://conferences.computer.org/icdar/2019/pdfs/ICDAR2019-5vPIU32iQjjaLtHlc8g8pO/7doDqZ3WIuLF2RHKwraE4k/1rPcoTBWZJldKF8sPWivxO.pdf}
}

@inproceedings{Li2020DocBank,
author    = {Minghao Li and Yiheng Xu and Lei Cui and Shaohan Huang and Furu Wei and Zhoujun Li and Ming Zhou},
title     = {DocBank: A Benchmark Dataset for Document Layout Analysis},
booktitle = {Proceedings of the 28th International Conference on Computational Linguistics},
series    = {COLING 2020},
year      = {2020},
month     = dec,
address   = {Barcelona, Spain (Online)},
publisher = {International Committee on Computational Linguistics},
pages     = {949--960},
doi       = {10.18653/v1/2020.coling-main.82},
url       = {https://aclanthology.org/2020.coling-main.82/}
}

@inproceedings{Pfitzmann2022DocLayNet,
author    = {Birgit Pfitzmann and Christoph Auer and Michele Dolfi and Ahmed S. Nassar and Peter Staar},
title     = {DocLayNet: A Large Human-Annotated Dataset for Document-Layout Segmentation},
booktitle = {Proceedings of the 28th ACM SIGKDD Conference on Knowledge Discovery and Data Mining},
series    = {KDD '22},
year      = {2022},
isbn      = {9781450393850},
pages     = {3743--3751},
numpages  = {9},
publisher = {Association for Computing Machinery},
address   = {New York, NY, USA},
doi       = {10.1145/3534678.3539043},
url       = {https://doi.org/10.1145/3534678.3539043}
}

@inproceedings{BirhaneForgottenMargins2022,
author = {Birhane, Abeba and Ruane, Elayne and Laurent, Thomas and Brown, Matthew S. and Flowers, Johnathan and Ventresque, Anthony and Dancy, Christopher L.},
title = {The Forgotten Margins of AI Ethics},
year = {2022},
isbn = {9781450393522},
publisher = {Association for Computing Machinery},
address = {New York, NY, USA},
url = {https://doi.org/10.1145/3531146.3533157},
doi = {10.1145/3531146.3533157},
booktitle = {Proceedings of the 2022 ACM Conference on Fairness, Accountability, and Transparency},
pages = {948--958},
numpages = {11},
location = {Seoul, Republic of Korea},
series = {FAccT '22}
}

@inproceedings{BirhaneValuesEncoded2022,
author = {Birhane, Abeba and Kalluri, Pratyusha and Card, Dallas and Agnew, William and Dotan, Ravit and Bao, Michelle},
title = {The Values Encoded in Machine Learning Research},
year = {2022},
isbn = {9781450393522},
publisher = {Association for Computing Machinery},
address = {New York, NY, USA},
url = {https://doi.org/10.1145/3531146.3533083},
doi = {10.1145/3531146.3533083},
booktitle = {Proceedings of the 2022 ACM Conference on Fairness, Accountability, and Transparency},
pages = {173--184},
numpages = {12},
location = {Seoul, Republic of Korea},
series = {FAccT '22}
}

@inproceedings{SmithTesseract2007,
  author={Smith, R.},
  booktitle={Ninth International Conference on Document Analysis and Recognition (ICDAR 2007)}, 
  title={An Overview of the Tesseract OCR Engine}, 
  year={2007},
  volume={2},
  number={},
  pages={629-633},
  doi={10.1109/ICDAR.2007.4376991}}

@inproceedings{Xu2021LayoutLMv2,
author    = {Yang Xu and Yiheng Xu and Tengchao Lv and Lei Cui and Furu Wei and Guoxin Wang and Yijuan Lu and Dinei Florencio and Cha Zhang and Wanxiang Che and Min Zhang and Lidong Zhou},
title     = {{LayoutLMv2}: Multi-Modal Pre-Training for Visually-Rich Document Understanding},
booktitle = {Proceedings of the 59th Annual Meeting of the Association for Computational Linguistics and the 11th International Joint Conference on Natural Language Processing},
series    = {ACL-IJCNLP 2021},
year      = {2021},
publisher = {Association for Computational Linguistics},
url       = {https://aclanthology.org/2021.acl-long.201/},
doi       = {10.18653/v1/2021.acl-long.201}
}

@article{Wei2024OCR20,
author  = {Hao Wei and Chao Liu and Jingye Chen and Jianfeng Wang and Lingpeng Kong and Yiheng Xu and Xiaohui Zhang},
title   = {General OCR Theory: Towards OCR-2.0 via a Unified End-to-End Model},
journal = {arXiv},
year    = {2024},
url     = {https://arxiv.org/abs/2409.01704}
}

@article{Bai2025Qwen25VL,
author  = {Shuai Bai and Kai Chen and Xiaohui Liu and Jianfeng Wang and Wenqi Ge and Shiqi Song and Jingdong Lin},
title   = {Qwen2.5-VL Technical Report},
journal = {arXiv},
year    = {2025},
url     = {https://arxiv.org/abs/2502.13923}
}

@article{Cui2025PaddleOCRVL,
author  = {Cheng Cui and Tianxiang Sun and Shiyao Liang and Tao Gao and Zhe Zhang and Jiajun Liu and Yu Ma},
title   = {PaddleOCR-VL: Boosting Multilingual Document Parsing via a 0.9B Ultra-Compact Vision-Language Model},
journal = {arXiv},
year    = {2025},
url     = {https://arxiv.org/abs/2510.14528}
}

@article{Poznanski2025olmOCR,
author  = {Adam Poznanski and others},
title   = {olmOCR: Scaling PDF Understanding with Internet Archive Data},
journal = {arXiv},
year    = {2025},
url     = {https://arxiv.org/abs/2502.18443}
}

@inproceedings{Kim2022Donut,
author = {Kim, Geewook and Hong, Teakgyu and Yim, Moonbin and Nam, JeongYeon and Park, Jinyoung and Yim, Jinyeong and Hwang, Wonseok and Yun, Sangdoo and Han, Dongyoon and Park, Seunghyun},
title = {OCR-Free Document Understanding Transformer},
year = {2022},
isbn = {978-3-031-19814-4},
publisher = {Springer-Verlag},
address = {Berlin, Heidelberg},
url = {https://doi.org/10.1007/978-3-031-19815-1_29},
doi = {10.1007/978-3-031-19815-1_29},
booktitle = {Computer Vision -- ECCV 2022: 17th European Conference, Tel Aviv, Israel, October 23--27, 2022, Proceedings, Part XXVIII},
pages = {498--517},
numpages = {20},
location = {Tel Aviv, Israel}
}

@article{Blecher2023Nougat,
author = {Blecher, Lukas and Neubig, Graham},
title = {Nougat: Neural Optical Understanding for Academic Documents},
journal = {arXiv},
year = {2023},
url = {https://arxiv.org/abs/2308.13418}
}

@inproceedings{Lewis2006IITCDIP,
author = {Lewis, David and Agam, Gady and Argamon, Shlomo and Frieder, Ophir and Grossman, David and Heard, Jonathan},
title = {Building a Test Collection for Complex Document Information Processing},
year = {2006},
isbn = {1595933697},
publisher = {Association for Computing Machinery},
address = {New York, NY, USA},
url = {https://doi.org/10.1145/1148170.1148307},
doi = {10.1145/1148170.1148307},
booktitle = {Proceedings of the 29th Annual International ACM SIGIR Conference on Research and Development in Information Retrieval},
pages = {665--666},
numpages = {2},
location = {Seattle, Washington, USA},
series = {SIGIR '06}
}

@inproceedings{VanLandeghem2023DUDE,
author = {Van Landeghem, Jordy and Powalski, Rafa{\l} and Tito, Rub{\`e}n and Jurkiewicz, Dawid and Blaschko, Matthew and Borchmann, {\L}ukasz and Coustaty, Micka{\"e}l and Moens, Sien and Pietruszka, Micha{\l} and Ackaert, Bertrand and Stanis{\l}awek, Tomasz and J{\'o}ziak, Pawe{\l} and Valveny, Ernest},
title = {Document Understanding Dataset and Evaluation (DUDE)},
year = {2023},
publisher = {IEEE},
booktitle = {Proceedings of the IEEE/CVF International Conference on Computer Vision},
pages = {19471--19483},
doi = {10.1109/ICCV51070.2023.01789}
}

@inproceedings{Borchmann2021DUE,
author = {Borchmann, {\L}ukasz and Pietruszka, Micha{\l} and Stanis{\l}awek, Tomasz and Jurkiewicz, Dawid and Turski, Micha{\l} and Szyndler, Karolina and Grali{\'n}ski, Filip},
title = {DUE: End-to-End Document Understanding Benchmark},
year = {2021},
booktitle = {Proceedings of the NeurIPS Datasets and Benchmarks Track},
publisher = {Neural Information Processing Systems Foundation},
url = {https://datasetsbenchmarksproceedings.neurips.cc/paper/2021/file/069059b7ef840f0c74a814ec9237b6ec-Paper.pdf}
}

@inproceedings{Heakl2025KITAB,
author = {Heakl, Ahmed and Sohail, Muhammad Abdullah and Ranjan, Mukul and Elbadry, Rania and Ahmad, Ghazi Shazan and El-Geish, Mohamed and Maher, Omar and Shen, Zhiqiang and Khan, Fahad Shahbaz and Khan, Salman},
title = {KITAB-Bench: A Comprehensive Multi-Domain Benchmark for Arabic OCR and Document Understanding},
year = {2025},
booktitle = {Findings of the Association for Computational Linguistics: ACL 2025},
publisher = {Association for Computational Linguistics},
pages = {22006--22024},
doi = {10.18653/v1/2025.findings-acl.1135}
}

@inproceedings{Lee2021Miseenpage,
   author = {Lee, B. C. G. and Ortiz Baco, J. and Salter, S. H. and Casey, J.},
   title = {Navigating the Mise-en-Page: Interpretive Machine Learning Approaches to the Visual Layouts of Multi-Ethnic Periodicals},
   booktitle = {CHR 2021: Computational Humanities Research Conference,},
   year = {2021},
   type = {Conference Proceedings}
}

@inproceedings{Lee2020NewspaperNavigator,
author = {Lee, Benjamin Charles Germain and Mears, Jaime and Jakeway, Eileen and Ferriter, Meghan and Adams, Chris and Yarasavage, Nathan and Thomas, Deborah and Zwaard, Kate and Weld, Daniel S.},
title = {The Newspaper Navigator Dataset: Extracting Headlines and Visual Content from 16 Million Historic Newspaper Pages in Chronicling America},
year = {2020},
publisher = {Association for Computing Machinery},
booktitle = {Proceedings of the 29th ACM International Conference on Information \& Knowledge Management},
pages = {3055--3062},
doi = {10.1145/3340531.3412767},
series = {CIKM '20}
}

@inproceedings{Dell2023AmericanStories,
author = {Dell, Melissa and Carlson, Jacob and Bryan, Tom and Silcock, Emily and Arora, Abhishek and Shen, Zejiang and D'Amico-Wong, Luca and Le, Quan and Querubin, Pablo and Heldring, Leander},
title = {American Stories: A Large-Scale Structured Text Dataset of Historical U.S. Newspapers},
year = {2023},
booktitle = {Proceedings of the NeurIPS Datasets and Benchmarks Track},
publisher = {Neural Information Processing Systems Foundation},
url = {https://proceedings.neurips.cc/paper_files/paper/2023/file/ffeb860479ccae44d84c0de32acd693d-Paper-Datasets_and_Benchmarks.pdf}
}

@misc{loc_chronicling_african_american,
  author       = {{Library of Congress}},
  title        = {Chronicling America: African American Newspapers},
  howpublished = {\url{https://www.loc.gov/collections/chronicling-america/titles/}},
  note         = {Filtered by subject ethnicity: African American; publications dated 1777--1963},
  year         = {2025},
  institution  = {Library of Congress}
}

@misc{howard_black_press_archives,
  author       = {{Howard University Moorland-Spingarn Research Center}},
  title        = {Black Press Archives},
  howpublished = {\url{https://msrc.howard.edu/black-press-archive}},
  note         = {Collection includes over 2,000 newspaper titles, 100,000 issues, and 2,847 microfilm reels},
  year         = {2025},
  institution  = {Howard University}
}

@article{burchardt_ocr_trustworthiness_2023,
  author    = {Burchardt, J{\o}rgen},
  title     = {Are Searches in {OCR}-generated Archives Trustworthy? An Analysis of Digital Newspaper Archives},
  journal   = {Jahrbuch f{\"u}r Wirtschaftsgeschichte / Economic History Yearbook},
  volume    = {64},
  number    = {1},
  pages     = {31--54},
  year      = {2023},
  doi       = {10.1515/jbwg-2023-0003},
  publisher = {De Gruyter}
}

@misc{crowley_digitizing_black_press,
  author       = {{The Crowley Company}},
  title        = {Preserving Voices: Digitizing Howard University’s Black Newspaper Collection},
  howpublished = {\url{https://thecrowleycompany.com/preserving-voices-digitizing-howard-universitys-black-newspaper-collection/}},
  note         = {Describes multi-year digitization efforts for the Black Press Archives},
  year         = {2024}
}

@misc{loc_ocr_reprocessing_2025,
  author       = {{Library of Congress}},
  title        = {Improving Search through OCR Reprocessing in Chronicling America},
  howpublished = {\url{https://blogs.loc.gov/headlinesandheroes/2025/04/ocr-reprocessing/}},
  year         = {2025},
  institution  = {Library of Congress}
}

@misc{nbc_howard_university,
  author       = {{NBC News via BET Staff}},
  title        = {Howard University Receives Grant To Archive Black Newspapers},
  howpublished = {\url{https://www.bet.com/article/u49dcv/howard-university-receives-grant-to-archive-black-newspapers}},
  note         = {Report on Howard University’s \$2M grant to digitize 60\% Black Press Archives, based on NBC News coverage},
  year         = {2022},
  month        = {Feb},
  organization = {BET / NBC News}
}

@article{Holley2009HowGC,
  title={How Good Can It Get? Analysing and Improving OCR Accuracy in Large Scale Historic Newspaper Digitisation Programs},
  author={Rose Holley},
  journal={D Lib Mag.},
  year={2009},
  volume={15},
  url={https://api.semanticscholar.org/CorpusID:2568435}
}

@inproceedings{Neudecker2021OCREvalSurvey,
author = {Neudecker, Clemens and Baierer, Konstantin and Gerber, Mike and Clausner, Christian and Antonacopoulos, Apostolos and Pletschacher, Stefan},
title = {A Survey of OCR Evaluation Tools and Metrics},
year = {2021},
isbn = {9781450386906},
publisher = {Association for Computing Machinery},
address = {New York, NY, USA},
booktitle = {Proceedings of the 6th International Workshop on Historical Document Imaging and Processing},
pages = {13--18},
numpages = {6},
location = {Lausanne, Switzerland},
series = {HIP '21},
doi = {10.1145/3476887.3476888},
url = {https://doi.org/10.1145/3476887.3476888}
}

@article{Reul2019OCR4all,
author = {Reul, Christian and Christ, Dennis and Hartelt, Alexander and Balbach, Nico and Wehner, Maximilian and Wick, Christoph and Grundig, Christine and Büttner, Andreas and Puppe, Frank},
title = {OCR4all—An Open-Source Tool Providing a (Semi-)Automatic OCR Workflow for Historical Printings},
journal = {Applied Sciences},
year = {2019},
volume = {9},
number = {22},
articleno = {4853},
url = {https://doi.org/10.3390/app9224853},
doi = {10.3390/app9224853}
}

@inproceedings{Rijhwani2020EndangeredOCRError,
author = {Rijhwani, Shruti and Anastasopoulos, Antonios and Neubig, Graham},
title = {OCR Post-Correction for Endangered Language Texts},
year = {2020},
booktitle = {Proceedings of the 2020 Conference on Empirical Methods in Natural Language Processing},
publisher = {Association for Computational Linguistics},
address = {Online},
pages = {5931--5942},
doi = {10.18653/v1/2020.emnlp-main.478},
url = {https://aclanthology.org/2020.emnlp-main.478/}
}

@article{Springmann2018,
author = {Springmann, Uwe and Reul, Christian and Dipper, Stefanie and Baiter, Johannes},
title = {Ground Truth for Training OCR Engines on Historical Documents in German Fraktur and Early Modern Latin},
journal = {arXiv},
year = {2018},
url = {https://arxiv.org/abs/1809.05501}
}

@inproceedings{Beshirov2022DuoSearch,
author = {Beshirov, Aleksandar and Hadzhieva, Siyana and Koychev, Ivan and Dobreva, Milena},
title = {DuoSearch: A Novel Search Engine for Bulgarian Historical Documents},
year = {2022},
booktitle = {Proceedings of the 44th European Conference on Information Retrieval},
series = {ECIR 2022},
publisher = {Springer},
address = {Cham, Switzerland},
pages = {265--269},
url = {https://doi.org/10.1007/978-3-030-99739-7_31},
doi = {10.1007/978-3-030-99739-7_31}
}

@misc{Gallon2016BlackDH,
author = {Gallon, Kim},
title = {Making a Case for the Black Digital Humanities},
year = {2016},
url = {https://dhdebates.gc.cuny.edu/read/untitled/section/fa10e2e1-0c3d-4519-a958-d823aac989eb},
note = {In *Debates in the Digital Humanities*},
publisher = {University of Minnesota Press}
}

@inproceedings{Smith2013InfectiousTexts,
author = {Smith, David A. and Cordell, Ryan and Dillon, Elizabeth Maddock},
title = {Infectious Texts: Modeling Text Reuse in Nineteenth-Century Newspapers},
booktitle = {Proceedings of the 2013 IEEE International Conference on Big Data},
year = {2013},
pages = {86--94},
publisher = {IEEE},
doi = {10.1109/BigData.2013.6691675}
}

@article{Casey2022BlackPress,
author = {Casey, John},
title = {{``We Need a Press---a Press of Our Own''}: The Black Press beyond Abolition},
journal = {Civil War History},
year = {2022},
volume = {68},
number = {2},
pages = {117--130},
doi = {10.1353/cwh.2022.0010}
}

@inproceedings{Smith2022ArchivalML,
author = {Taurino, Giulia and Smith, David A.},
title = {Machine Learning as an Archival Science: Narratives behind Artificial Intelligence, Cultural Data, and Archival Remediation},
booktitle = {NeurIPS Workshop on AI Cultures},
year = {2022},
url = {https://ai-cultures.github.io/papers/machine_learning_as_an_archiva.pdf}
}

@inproceedings{Levchenko2024HistoricalCorpora,
author = {Levchenko, Maria},
title = {Building Historical Corpora with Multimodal {LLM}s: Epistemic Gaps and Misreadings in 18th-Century Russian Books},
booktitle = {ACH Anthology},
year = {2024},
url = {https://doi.org/10.63744/SKoZVUHQbtE7/},
publisher = {Association for Computers and the Humanities},
address = {United States}
}

@article{Page2021PRISMA,
author = {Page, Matthew J. and McKenzie, Joanne E. and Bossuyt, Patrick M. and Boutron, Isabelle and Hoffmann, Tammy C. and Mulrow, Cynthia D. and Shamseer, Larissa and Tetzlaff, Jennifer M. and Akl, Elie A. and Brennan, Sue E. and Chou, Roger and Glanville, Julie and Grimshaw, Jeremy M. and Hróbjartsson, Asbjørn and Lalu, Manoj M. and Li, Tianjing and Loder, Elizabeth W. and Mayo-Wilson, Evan and McDonald, Steve and McGuinness, Luke A. and Stewart, Lesley A. and Thomas, James and Tricco, Andrea C. and Welch, Vivian A. and Whiting, Penny and Moher, David},
title = {The {PRISMA} 2020 Statement: An Updated Guideline for Reporting Systematic Reviews},
journal = {BMJ},
year = {2021},
volume = {372},
articleno = {n71},
url = {https://www.bmj.com/content/372/bmj.n71},
doi = {10.1136/bmj.n71}
}

@book{Noble2018,
   author = {Noble, S. U.},
   title = {Algorithms of oppression: How search engines reinforce racism},
   publisher = {NYU Press},
   address = {New York, NY, USA},
   ISBN = {1479837245},
   year = {2018},
   type = {Book}
}

@article{Dancy2022AIBlackness,
   author = {Dancy, C. L. and Saucier, P. K.},
   title = {AI and Blackness: Towards moving beyond bias and representation},
   journal = {IEEE Transactions on Technology and Society},
   volume = {3},
   number = {1},
   pages = {31-40},
   ISSN = {2637-6415},
   DOI = {10.1109/TTS.2021.3125998},
   year = {2022},
   type = {Journal Article}
}

@inproceedings{Croskey2025LibCollections,
author = {Croskey, Payton and Offert, Fabian and Jacobs, Jennifer and Thaler, Kai M.},
title = {Liberatory Collections and Ethical AI: Reimagining AI Development from Black Community Archives and Datasets},
year = {2025},
isbn = {9798400714825},
publisher = {Association for Computing Machinery},
address = {New York, NY, USA},
url = {https://doi.org/10.1145/3715275.3732058},
doi = {10.1145/3715275.3732058},
booktitle = {Proceedings of the 2025 ACM Conference on Fairness, Accountability, and Transparency},
pages = {900--913},
numpages = {14},

series = {FAccT '25}
}

@article{Morrison2022BlackComputationalThought,
   author = {Morrison, R.},
   title = {Voluptuous Disintegration: A Future History of Black Computational Thought},
   journal = {DHQ: Digital Humanities Quarterly},
   volume = {16},
   number = {3},
   ISSN = {1938-4122},
   year = {2022},
   type = {Journal Article}
}

@book{Taylor2019RaceProfit,
   author = {Taylor, K.-Y.},
   title = {Race for profit: How banks and the real estate industry undermined black homeownership},
   publisher = {UNC Press},
   address = {Chapel Hill, NC},
   ISBN = {1469653672},
   year = {2019},
   type = {Book}
}

\appendix
\section{Appendix A: Search Queries Used for PRISMA 2020 Review}

\begin{table}[h]
\raggedright
\caption{Search queries used in the PRISMA 2020 systematic review of OCR evaluation literature.}
\begin{tabular}{p{0.35\linewidth} p{0.55\linewidth}}
\hline
\textbf{Query Category} & \textbf{Search Query} \\
\hline
Primary evaluation queries
& ``OCR evaluation metrics document understanding'' \\
& ``document AI benchmark evaluation'' \\
& ``OCR benchmark dataset peer reviewed'' \\
& ``document layout analysis benchmark'' \\
& ``vision language model OCR evaluation'' \\
\hline
Historical \& bias-aware queries
& ``historical document OCR evaluation'' \\
& ``OCR bias historical documents'' \\
& ``newspaper digitization OCR evaluation'' \\
& ``layout preservation OCR evaluation'' \\
& ``cultural heritage document analysis OCR'' \\
\hline
Model-class specific queries
& ``vision transformer OCR evaluation'' \\
& ``multimodal document understanding benchmark'' \\
& ``end-to-end document understanding evaluation'' \\
\hline
Exclusion / sanity-check queries
& ``synthetic document OCR benchmark'' \\
& ``handwritten text recognition evaluation dataset'' \\
\hline
\end{tabular}
\raggedright
\end{table}

\end{document}